%% file: main_.tex
\documentclass[letterpaper, 10 pt, conference]{ieeeconf}  
\IEEEoverridecommandlockouts   
\usepackage{amsfonts}
\usepackage{amsthm}
\makeatletter
\def\th@plain{%
  \thm@notefont{}
  \itshape 
}

\makeatother
\usepackage{mathtools}      
\usepackage{amssymb}

\usepackage[linesnumbered,algoruled,boxed]{algorithm2e}
\usepackage{dsfont}
\usepackage{comment}
\usepackage{xcolor}
\usepackage{graphicx}
\usepackage{cite}
\usepackage[font=small,labelfont=bf]{caption}
\usepackage{subcaption}
\usepackage{diagbox}
\usepackage{algorithmic}
\usepackage{wrapfig}
\usepackage{hyperref}
\let\labelindent\relax
\usepackage{enumitem}

\usepackage{overpic}        
\usepackage{tikz}

\usepackage[textsize=footnotesize]{todonotes}
\usepackage{xcolor}
\usepackage{xargs} 
\newcommandx{\kg}[2][1=]{\todo[linecolor=red,
			backgroundcolor=red!10,bordercolor=red,#1]{#2}}
\newcommandx{\jy}[2][1=]{\todo[linecolor=green,
			backgroundcolor=green!10,bordercolor=green,#1]{JY: #2}}
\newcommandx{\sw}[2][1=]{\todo[linecolor=blue,
			backgroundcolor=blue!10,bordercolor=blue,#1]{SW: #2}}

\usepackage[normalem]{ulem}

\newtheoremstyle{mystyle}
  {}
  {}
  {\itshape}
  {}
  {\bfseries}
  {.}
  { }
  {}

\theoremstyle{mystyle}
\newtheorem{problem}{Problem}[section]

\theoremstyle{definition}
\newtheorem{definition}{Definition}[section]
\theoremstyle{remark}
\newtheorem*{remark}{Remark}

\def\prob{{\texttt{{BDHD}}}\xspace}
\def\ours{{{{EDP}}}\xspace}

\SetKwProg{Fn}{Function}{}{}
\SetKwComment{Comment}{$\%\%$\ }{}
\SetKw{Continue}{continue}

\font\titlefont=ptmb at 16pt
\title{\titlefont
Efficient Algorithms for Boundary Defense with Heterogeneous Defenders 
}

\author{ Si Wei Feng \and Jingjin Yu
\thanks{ S. W. Feng, and J. Yu are with the Department of Computer Science, Rutgers, the State University of New Jersey, Piscataway, NJ, 
 USA. E-Mails: \{{\tt siwei.feng,  jingjin.yu}\}\hspace*{.25em}
 @ \hspace*{.25em}rutgers.edu.
}
\thanks{This work is partly supported by NSF award IIS-1845888 and an Amazon Research Award.
}
}

\begin{document}
\maketitle
\thispagestyle{empty}
\pagestyle{empty}
\nocite{adler2022role}
\begin{abstract}
This paper studies the problem of defending (1D and 2D) boundaries against a large number of continuous attacks with a heterogeneous group of defenders. 
The defender team has perfect information of the attack events within some time (finite or infinite) horizon, with the goal of intercepting as many attacks as possible. An attack is considered successfully intercepted if a defender is present at the boundary location when and where the attack happens. 
Through proposing a network-flow and integer programming-based method for computing optimal solutions, 
and an exhaustive defender pairing heuristic method for computing near-optimal solutions, 
We can significantly reduce the computation burden in solving the problem compared to the previous state of the art. 
Extensive simulation experiments confirm the effectiveness of the algorithms. 
Leveraging our efficient methods, we also characterize the solution structures, revealing the relationships between the attack interception rate and the various problem parameters, e.g., the heterogeneity of the defenders, attack rate, boundary topology, and the look-ahead horizon. 

\vspace{1mm}
\noindent Introduction video: \url{https://youtu.be/x0mQD\_7RhKI}.
\end{abstract}

\section{Introduction}

\input{tex/intro}\label{sec:intro}

\section{Preliminaries}\label{sec:preliminary}
\input{tex/problem}

\section{Algorithmic Solutions for \prob}\label{sec:algorithm}
\input{tex/algo}

\section{Evaluation and Empirical Study of \prob}\label{sec:evaluation}
\input{tex/experiments}


\section{Conclusion}\label{sec:conclusion}

\input{tex/conclusion}

\bibliography{bib/bib}
\bibliographystyle{IEEEtran}

\end{document}

%% file: tex/intro.tex
Perimeter-defense problems model scenarios where the interior of a region must be defended against a set of incoming attacks.
With a team of defenders (i.e., mobile robots and/or agents) located on the perimeter of a given region, an attack is considered successfully intercepted if at least one defender is present at the attack location when the attack happens. 
This problem finds broad applications, including the protection of endangered wildlife \cite{haksar2020spatial}, aerial defense \cite{lykou2020defending, lee2020perimeter}, and border security \cite{agmon2008multi, fengoptimally}, to list a few. 

The study of perimeter defense problems finds some of its origins in target guarding problems \cite{rufus1965}, a classical pursuit-evasion game studying the strategies of multiple defenders to guard a static region.
In \cite{shishika2018local}, the target-guarding problem was first specialized to produce perimeter defense in which the perimeter is the target to be guarded, and defenders can only move on the perimeter. 
Various methods have been developed, including region decomposition \cite{shishika2018local}, assignment or matching-based algorithm \cite{shishika2020cooperative}, network flow formulation \cite{chen2021optimal}, and coverage control-based method \cite{macharet2020adaptive}.

Perimeter-defense problems may also be considered a variant of the reach-avoid game between two parties, where one party tries to send as many attackers to a region the other party must defend \cite{rufus1965}. 
Studies of the reach-avoid game usually apply specific case analysis, which suffers from the exponential increase in time complexity and size of state space as the number of attackers increase \cite{margellos2011hamilton, zhou2012general, yan2018reach}. 
Recent work leans toward adopting maximum matching or assignment frameworks from graph theory to address this problem for multiple defenders \cite{chen2014path, chen2014multiplayer, yan2019matching}.


In this work, we study perimeter-defense problems under a perfect information assumption that the attackers' attack time and locations on the perimeter are known as soon as attackers appear. The assumption is adopted in several recent related research including \cite{adler2022role, macharet2020adaptive}.
The particular emphasis of our work is on the more challenging case of heterogeneous defenders, where the defenders have different speeds in responding to incoming attacks. 
We denote the problem as \emph{boundary-defense with heterogeneous defenders} or \prob.\footnote{We use \emph{boundary} instead of \emph{perimeter} since perimeter generally refers to the 1D boundary of a 2D geometry shape. 2D hemisphere has been examined in \cite{lee2020perimeter} for constraining defender.}
The heterogeneous setting for perimeters is first carefully investigated in \cite{adler2022role}, which carried out the detailed structural study and introduced an algorithm based on dynamic programming (DP) \cite{cormen2022introduction} for handling infinite horizon attacking sequences, focusing mostly on two defenders. 

\begin{figure}[t]
    \centering
    \includegraphics[width=1\linewidth]{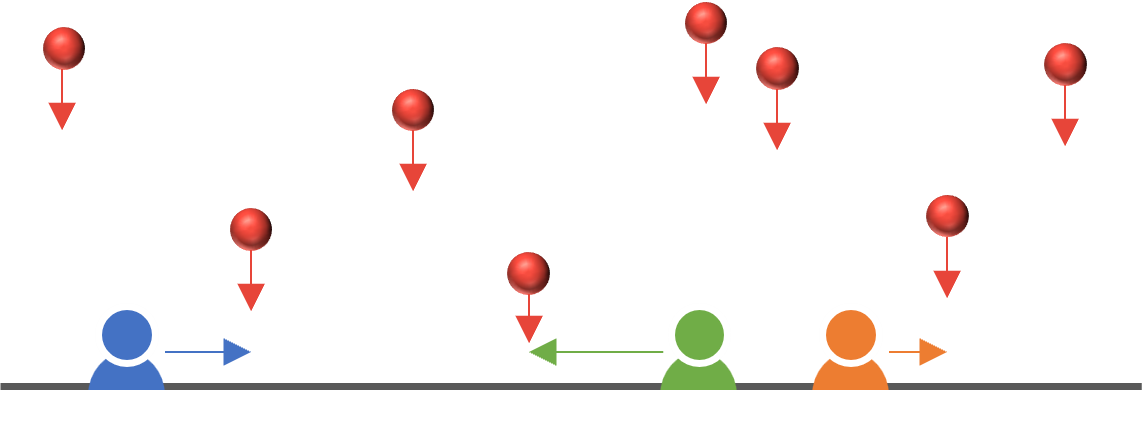}
    \vspace{-6mm}
    \caption{Illustration of the problem of boundary defense with heterogeneous defenders, where multiple defenders with different capabilities must do their best to intercept incoming attacks (signified as red balls moving downwards). The defenders are constrained to move on the boundary, which is a 1D perimeter in this case.}
    \label{fig:bdhd}
    \vspace{-5mm}
\end{figure}


The main contribution of our work lies with the development of efficient computational tools for and an empirical structural study of \prob. More specifically, 
\begin{itemize}[leftmargin=3.5mm]
\item We developed a network-flow formulation of \prob, leading to an exact method, based on integer linear programming (ILP), for solving the problem. The IP-based method is much more scalable than the DP-based method (which is only effective for no more than three defenders). 
\item Using the DP solution as a sub-routine, we developed a highly scalable heuristics method, called \emph{exhaustive defender pairing} (\ours), that runs in low polynomial time.  \ours is demonstrated to compute high-quality (in fact, often optimal or near-optimal) solutions. The design philosophy of \ours is of independent algorithmic interest. 
\item We further generalize our algorithms to apply to finite-horizon \prob (more challenging than infinite-horizon \prob), where only attacks whose attack time is within some $T$ time of hitting the boundary are revealed.
\item Utilizing the scalable algorithms developed in this work, we performed a thorough empirical study of the solution characteristics of \prob under various possible problem settings, including (1) different attack densities, (2) different defender speed distributions, and (3) different domain topology, i.e., $S^1$ (circle), $I = [0, 1]$ (unit interval), $S^2$ (sphere), and $I \times I$ (unit square). 
\end{itemize}


The rest of the paper is organized as follows. 
In Sec.~\ref{sec:preliminary}, we formulate the problems studied in this paper and introduce the notations used. 
In Sec.~\ref{sec:algorithm}, we describe the previous dynamic programming method in \cite{adler2022role}, and our ILP-based algorithms and \ours. 
We discuss the performance of our algorithms and solution characteristics of \prob under different settings in Sec.~\ref{sec:evaluation}, and conclude with Sec.~\ref{sec:conclusion}.

%% file: tex/problem.tex
In boundary defense with heterogeneous defenders (\prob), there are $k$ defenders (which may be robots and/or other types of agents) with speeds $v_1,\dots,v_k$, where each defender is modeled as a point in some domain $\mathcal E = \mathbb R^2$ or $\mathbb R^3$.
The defenders live on a lower dimensional subspace of $\mathcal E$ (i.e., some boundary of a subset of $\mathcal E$).  
There are also $n$ attack events $\big\langle loc_i, t_i\big\rangle_{i=1}^{n}$, where each attack event is a pair $\big\langle loc, t\big\rangle$ in which $loc$ is the
location of the attack and $t$ is the time it happens. 
The $i^{{th}}$ attack is intercepted by a defender only if the defender is located at $loc_i$ at time $t_i$.
For initialization, we denote the initial locations of the $k$ defenders at $t=0$ as $loc^{1},\dots, loc^{k}$. 

Following the definition in \cite{adler2022role}, we define the horizon of the defenders as follows,
\begin{definition}[Horizon]
The (look ahead) \textit{horizon} $T$ of the defenders is defined as the amount of time defenders can peek into the attack sequence in the future. That is, given the current time $t$, and a horizon $T$, defenders have access to complete information on attacks happening on or before $t+T$. 
\end{definition}

Now, we provide formulations of the two versions of the \prob problem studied in this paper. In the infinite horizon setting, $T = \infty$, all attack events are given in a single batch.

\begin{problem}[Infinite-horizon \prob]\label{prob:1}
Given $k$ defenders with speed $v_1, \dots, v_k$ and initial locations $loc^1, \ldots, loc^k$, and $n$ attack events $\big\langle loc_i, t_i\big\rangle_{i=1}^{n}$, intercept as many attacks as possible. 
\end{problem}

In a finite-horizon setting, the attack events are not all revealed at $t=0$ but are given as a stream of attacks $\big\langle loc_i, t_i\big\rangle_{i=1}^{\infty}$. The defenders can only know the attack events within a horizon or time window of $T < \infty$ in the future.

\begin{problem}[Finite-horizon \prob]\label{prob:2}
There are $k$ defenders with speed $v_1, \dots, v_k$ and initial locations $loc^1, \ldots, loc^k$, and a stream of attack events.
At each time instance $t$, the defenders only know attack events happening no later than time $T$ in the future (i.e. $t_i \le t+T$).
Intercept as many attacks as possible. 
\end{problem}

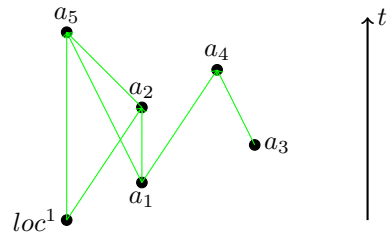
\begin{figure}[h]
    \centering
\begin{tikzpicture}
\draw[black, thick, ->] (5,0) -- (5, 2.7) node[anchor=west]{$t$};

\filldraw[black] (1, 0) circle (2pt) node[anchor=east](iloc1){$loc^1$};

\filldraw[black] (2,0.5) circle (2pt) node[anchor=north](a1){$a_1$};
\filldraw[black] (2,1.5) circle (2pt) node[anchor=south](a2){$a_2$};
\filldraw[black] (3.5,1) circle (2pt) node[anchor=west](a3){$a_3$};
\filldraw[black] (3,2) circle (2pt) node[anchor=south](a4){$a_4$};
\filldraw[black] (1,2.5) circle (2pt) node[anchor=south](a5){$a_5$};

\draw[green, thin, ->] (iloc1.east) -- (a2.south);
\draw[green, thin, ->] (iloc1.east) -- (a5.south);

\draw[green, thin, ->] (a1.north) -- (a2.south);
\draw[green, thin, ->] (a1.north) -- (a4.south);
\draw[green, thin, ->] (a1.north) -- (a5.south);
\draw[green, thin, ->] (a2.south) -- (a5.south);
\draw[green, thin, ->] (a3.west) -- (a4.south);

\end{tikzpicture}
    \caption{Illustration of an example of reachability between different attack events for defender $1$.  
    For example, $next(1, 1) = \{2,4,5\}$, $next(0,1)=\{2,5\}$, $prev(2, 1) =\{0, 1\}$, $prev(1,1)=\varnothing$.
    $loc^1$ cannot reach $a_1$ since $v_1$ is not sufficiently large. For the same reason, defender $1$ cannot reach $a_3$ from $a_1$.
    }
    \label{fig:next_prev}
\end{figure}

We introduce here two useful notations: $next(a, d)\ (a\in[0, n],\ d\in[1,k])$ and $prev(a, d)\ (a\in[1,n],\ d \in [1,k])$.
They are defined as
\begin{align}
next(a, d) = \{ a'| dist(loc_a, loc_{a'}) \leq v_d \cdot (t_{a'} - t_a) \}    \\
prev(a, d) = \{ a'| dist(loc_a, loc_{a'}) \leq v_d \cdot (t_a - t_{a'}) \}    
\end{align}
where $dist(x,y)$ denotes the distance between location $x$ and $y$. 
In other words, $next(a, d)$ is the set of 
attack events that can be reached from the location of the $a^{th}$ attack event by defender $d$. 
And $prev(a,d)$ is the set of attack events from whose location defender $d$ can reach the $a^{th}$ attack event. 
Additionally, $next(0, d)$ denotes the set of attack events that can be reached by defender $d$ from its initial location.
Similarly, $prev(a, d)$ contains $0$ if defender $d$ can reach the location of attack $a$ from its initial location.
Fig.~\ref{fig:next_prev} gives an example for defender $1$.

%% file: tex/algo.tex
In this section, we describe three methods for solving infinite-horizon \prob (Problem~\ref{prob:1}).
Sec.~\ref{sec:dp} provide a dynamic programming (DP) algorithm based on one developed in \cite{adler2022role}, 
followed by Sec.~\ref{sec:ilp} formulating an integer linear programming model,
and Sec.~\ref{sec:dp_local} discusses a heuristic search algorithm based on the DP algorithm. 
The extension to an infinite attack stream with a finite look-ahead horizon  (Problem~\ref{prob:2}) will be discussed in Sec.~\ref{sec:hor}.

\subsection{Exact Dynamic Programming Based Method}
\label{sec:dp}

The DP algorithm builds a recursion formula on the last attacks intercepted by the defenders.
Without loss of generality, we assume that each attack is only intercepted by one defender.
For a DP state $a_1, \dots, a_k$ where $a_i$ is the last attack defender $i$ intercepts, 
let $T[a_1]\dots[a_k]$ store the maximum number of attacks that can be intercepted when the $i^{th}$ defender's last intercepted attack events is $a_i$ $(i\in[1,k])$,
and denote $a_{ma}$ as the maximum of $a_1, \dots, a_k$.
Base on which attack the $ma^{th}$ defender intercepts before intercepting attack $a_{ma}$,
we can write the DP recursion formula as follows.
\begin{equation}
\begin{split}
T[a_1]\dots[a_{ma}]\dots[a_k] &= \\ 
\max_{p\in prev(a_{ma}, ma) \wedge p\neq a_1 \dots a_k} &  T[a_1]\dots[p]\dots[a_k] + 1
\end{split}
\end{equation}

Pseudo-code in Alg.~\ref{alg:dp} provides a sketch of a possible implementation of the dynamic programming algorithm.
Effectively implemented, the time complexity of running Alg.~\ref{alg:dp} is $O( (n+1)^{k+1})$, which is polynomial when $k$, the number of defenders, is fixed.
\vspace{-2mm}

\begin{algorithm}
\DontPrintSemicolon
\KwData{
$E=\big \langle loc_i, t_i\big\rangle_{i=1}^{n}$: $n$ attack events\;
$loc^1,\dots,loc^k$: initial locations of the $k$ defenders\;
$v_1,\dots,v_k$: speeds of the $k$ defenders\;
}
\KwResult{Maximum number of attacks intercepted}
\vspace{1mm}
$T\gets$ an $(n+1)^k$-length array initialized to $-\infty$\;
\vspace{1mm}
$result\gets 0$\;
\vspace{1mm}
$T[0]\gets 0$\;
\vspace{1mm}
\For{$mask\gets 0 $ \KwTo $(n+1)^k - 1$}{
\vspace{1mm}
    $\overline{a_1 a_2\dots a_k} \leftarrow mask$\;
    \Comment{$\overline{a_1 a_2\dots a_k}$ represents a base-($n+1$) number, i.e., $a_1\cdot (n+1)^{k-1} + a_2 \cdot (n+1)^{k-2} +\dots+ a_k$}
    \If{$\exists\ a_i = a_j$}{
        \Continue\;
    }
\vspace{1mm}
    $ma \leftarrow argmax_i a_i$\;
\vspace{1mm}
    \For{$p\in prev(a_{ma}, ma)$}{
        \If{$\forall i\ pos_i \neq p$}{
            $pm\gets mask - (a_{ma} - p)\cdot (n+1)^{k-ma}$\;
            \Comment{$pm$ is the result of replacing $a_{ma}$ with $p$}
            $T[mask]\gets \max(T[mask], T[pm] + 1)$\;
        }
    }
\vspace{1mm}
    $result\gets \max(result, T[mask])$\;
}
\vspace{1mm}
\Return{$result$}\;

\caption{Dynamic Programming for \prob}
\label{alg:dp}
\end{algorithm}
\vspace{-2mm}

The algorithm presented here is a slight modification of the DP algorithm in \cite{adler2022role} with two subtle differences. First, we enforce that an attack event can only be handled by one defender.
Second, we explicitly use the initial locations of the defenders in the algorithm, which is essential for handling the finite horizon extension.

\subsection{Solving \prob with Integer Linear Programming Model based on a Flow Formulation}
\label{sec:ilp}

It is not difficult to see that \prob can also be seen as a network flow problem by treating each attack event as a node and the reachability between each pair of attack events for each defender as the edges in the graph. 

Specifically, there are $n$ nodes in the graph, representing the $n$ attack events. 
There are also $O(n^2k)$ connections between nodes inside the graph.
If defender $j$ can reach attack event $i'$ from $i$, there is an connection 
$edge[i][i'][j]$ between node $i$ and $i'$. 
Also, we use a binary variable $intercept[i]$ to denote whether attack event $i$ is successfully intercepted. 
These give rise to the following integer linear programming (ILP) formulation of the problem.

Eq. \eqref{eq:intercept} sets the criteria of an attack $i$ being intercepted as at least one defender to come into the node $i$, which means intercept attack $i$.
Eq. \eqref{eq:flow} sets the defender flow conservation rule that the number of type $j$ defender exiting node $i$ must be larger than or equal to the number of type $j$ defender coming to node $i$.
Eq. \eqref{eq:initial} sets the initial constraints on the number of each type of defender used (coming out from node 0).

\begin{gather}
\sum_{j\in[1, k],\ i'\in prev(i, j)} edge[i'][i][j] \geq intercept[i] \label{eq:intercept}\\
\sum_{nxt_i\in next(i, j)} edge[i][nxt_i][j] \leq \sum_{prv_i \in prev(i,j)} edge[prv_i][i][j]  \label{eq:flow} \\
\sum_{nxt_0 \in next(0, j)} edge[0[nxt_0][j] \leq 1 \label{eq:initial}  \\
Objective\quad \max \sum_i intercept[i]
\end{gather}

Denote $M$ as the number of connections in the graph, clearly $M<n^2k$.
This integer linear programming formulation uses $M + n$ variables, and 
$nk + n$ constraints. 

\begin{remark}
The flow formulation of the problem is an NP-hard problem. The proof is similar to the NP-completeness proof of the two-commodity flow in \cite{even1975complexity}. This seems to suggest that \prob is NP-hard as well. 
\end{remark}


\subsection{Exhaustive Defenders Pairing Heuristic Search Method}
\label{sec:dp_local}
We now develop a heuristic search method using the dynamic programming algorithm discussed in Alg.~\ref{alg:dp} applied on two defenders.
The DP algorithm that computes the optimal solution for two defenders is used as a local improvement primitive for the local search heuristic algorithm. 

We call the resulting algorithm \emph{exhaustive defender pairing} (\ours).  
In each iteration, \ours pick two defenders, and the attack events the two defenders have intercepted and the attack events that have not been intercepted by any defender. 
Then, \ours uses the DP algorithm in Alg.~\ref{alg:dp} to increase the number of attack events intercepted by the two defenders selected. 
The complete algorithm is sketched in Alg.~\ref{alg:dp_local}.

\begin{algorithm}[h]
\DontPrintSemicolon
\KwData{
$E=\big \langle loc_i, t_i\big\rangle_{i=1}^{n}$: $n$ attack events\;
$loc^1,\dots,loc^k$: initial locations of the $k$ defenders\;
$v_1,\dots,v_k$: speeds of the $k$ defenders\;
}
\KwResult{Number of attacks intercepted}

$Intercept \gets$ a length-$n$ array initialized to $-1$\;
\Comment{$Intercept$ array stores for each event the defender that intercepts it}
$result\gets 0$\;
\vspace{1mm}
\For{$u, v \in \{1, \dots, k\}\times\{1, \dots, k\}, u\neq v$}{
    $E'\gets \{w\ |\ Intercept[w] \in\{u, v, -1\}\}$ \;
    \Comment{$E'$ stores the set of attack events intercepted by defender $u,v$ and the attacks not intercepted by any defender}
\vspace{1mm}
    $\tilde{n}\gets E'.size$\;
\vspace{1mm}
    T $\gets$ an $(\tilde{n}+1)^2$-length array initialized to $-\infty$\;
\vspace{1mm}
    $T[0]\gets 0$\;
    \Comment{Apply the DP algorithm for defender $u$ and $v$}
\vspace{1mm}
    \For{$mask\gets 0 $ \KwTo $(\tilde{n}+1)^2-1$}{
        $\overline{a_1 a_2} \leftarrow mask$\;
        \If{$a_1 =a_2$}{
            \Continue\;
        }
\vspace{1mm}
        $ma \leftarrow argmax_i a_i$\;
\vspace{1mm}
        \For{$p\in prev(a_{ma}, ma)$}{
            \If{$\forall i\ a_i \neq p$}{
                $pm\gets mask - (a_{ma} - p)\cdot (n+1)^{2-ma}$\;
                $T[mask]\gets \max(T[mask], T[pm] + 1)$\;
            }
        }
    }
\vspace{1mm}
    \If{Solution is improved}{
        Update $Intercept, result$\;
    }
}
\vspace{1mm}
\Return{$result$}
\caption{Exhaustive Defender Pairing}
\label{alg:dp_local}
\end{algorithm}

We can try different defender pairing orders and choose the best one. 
In our \ours implementation, we choose to run line 3 of Alg.~\ref{alg:dp_local} in 3 different iteration ordering of $u, v$.
Alg.~\ref{alg:dp_local}'s running time is $O(k^2 n^3)$ as we try $O(k^2)$ pairs of defenders, and each run of the DP algorithm takes $O(n^3)$.

\subsection{Handling Infinite Attack Streams with a Finite Look-Ahead Horizon }
\label{sec:hor}
For Problem~\ref{prob:2} where the attacks $\big \langle loc_i, t_i\big \rangle_{i=1}^{\infty}$ may be infinite, and the look-ahead horizon is finite, not all attacks are revealed at once. As such, previous methods cannot be directly applied. 
As the future attack sequence cannot be foreseen, defenders need to \emph{react} to information (e.g., the attack events in the next $T$ time interval) obtained so far. 

Towards addressing the problem, a greedy \emph{replanning} approach can be applied. Whenever an attack event is observed, the defender team replans the capture sequence given the new attacks added to the attack queue. 
This gives rise to an online algorithm sketched in Alg.~\ref{alg:horizon}.

\begin{algorithm}[h]
\DontPrintSemicolon
\KwData{
$E=\big \langle loc_i, t_i\big\rangle_{i=1}^{\infty}$: a stream of attack events\;
$loc^1,\dots,loc^k$: initial locations of the $k$ defenders\;
$v_1,\dots,v_k$: speeds of the $k$ defenders\;
}
$E'\gets $ an empty queue\;
\Comment{$E'$ stores attack events seen so far}
\vspace{1mm}
\While{new attack events added to $E'$ }{
    Apply Alg.~\ref{alg:dp_local} to compute a plan for the defenders and attack events $E'$\;
    Execute the plan, pop out from $E'$ attack events passed, and update $loc^1,\dots,loc^k$ to the defenders' current locations until new attacks are foreseen\;
}

\caption{Online Exhaustive Defender Pairing}
\label{alg:horizon}
\end{algorithm}

%% file: tex/experiments.tex
In this section, we describe our experimental study of the methods mentioned or proposed in the paper, which serves two purposes: (1) demonstrating the performance of our fast computational methods, and (2) characterizing the solution structure of \prob under different problem settings.

The algorithms are implemented in C++ and run on a 3.6GHz quad-core CPU with 16G memory.
For integer linear programming, Gurobi \cite{optimization2019gurobi} is used as the solver with a time limit of 60 seconds applied.
In the evaluation, two main factors are considered: computation time and solution quality measured by interception rate. 

The instances were generated for up to dozens of defenders, 
and the defender speeds are sampled uniformly at random from $1$ to $v_{max}$.
The attack sequence was generated according to a Poisson process with parameter $\lambda$ (i.e., the time gap between
two consecutive attacks conforms with an exponential distribution $P(t) = \lambda e^{-\lambda t}$),
and the locations of the attack are sampled uniformly at random from the boundary.
We use 4 types of boundary for testing a) unit circle $S^1$, b) unit interval $I=[0, 1]$
c) unit square $I^2=[0, 1]\times[0,1]$, and d) unit sphere $S^2$.

\subsection{Infinite-Horizon \prob: Basic Performance Evaluation}
This section provides experimental study and comparisons between the three methods described in the paper: 
dynamic programming (DP), integer linear programming (ILP), and exhaustive defender pairing (\ours) algorithms.

\begin{figure}[h!]
    \centering
    \includegraphics[width=\linewidth]{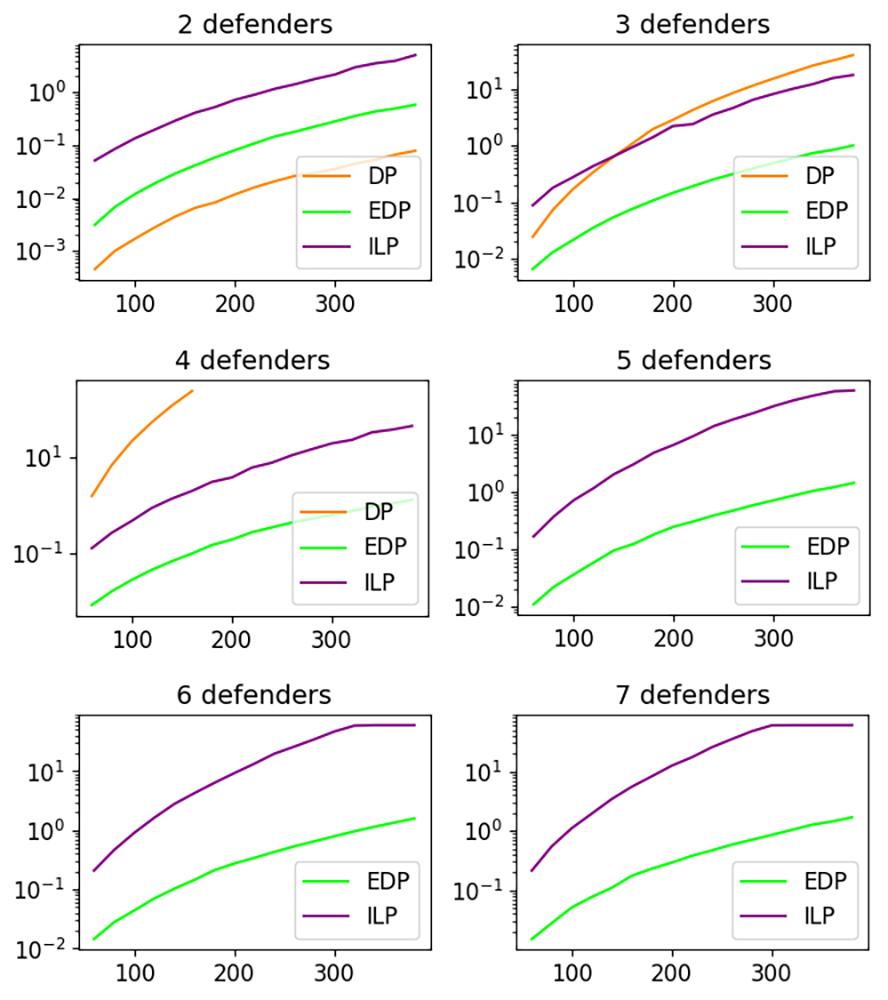}
    \caption{Computation time in seconds ($y$-axis) for dynamic programming (DP), integer linear programming (ILP) and \ours for 2 to 7 defenders and up to 400 attack events ($x$-axis). DP quickly becomes intractable as the number of defenders goes beyond $3$; ILP is about 2-3 magnitudes slower than \ours.}
    \label{fig:time_cost}
\end{figure}

\begin{figure}[h!]
    \centering
    \includegraphics[width=\linewidth]{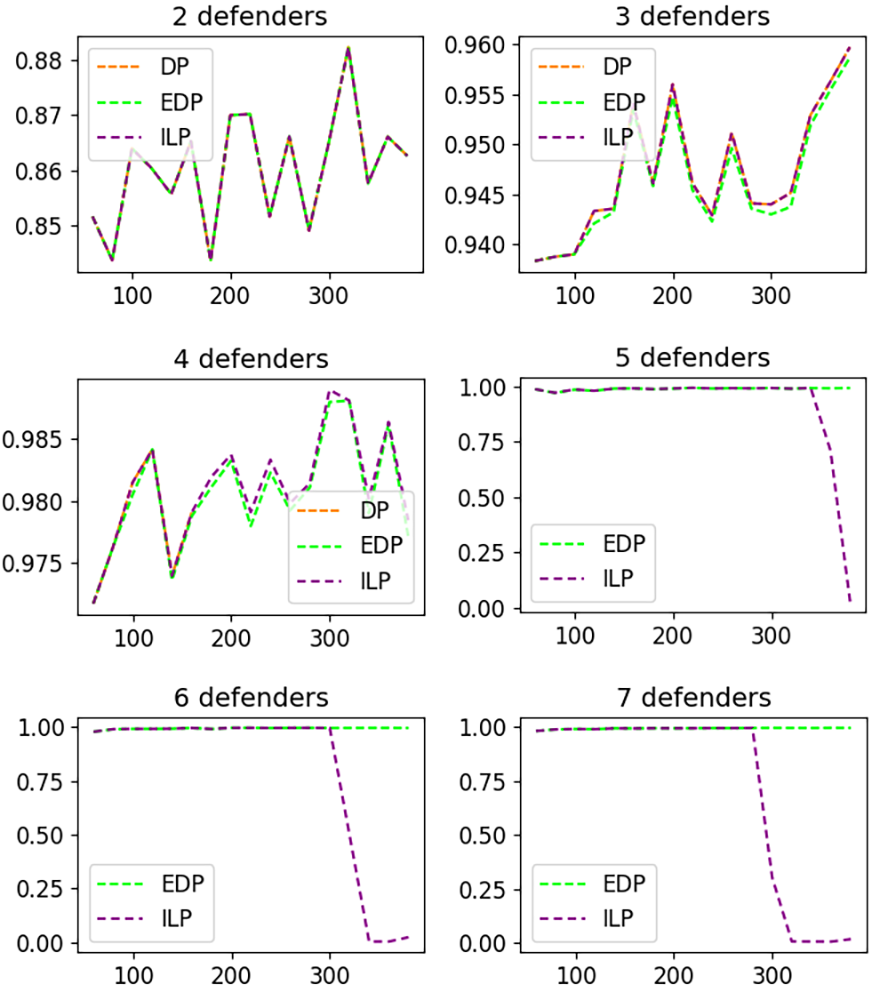}
    \caption{Solution quality (interception rate) for dynamic programming (DP), integer linear programming (ILP) and \ours  for 2 to 7 defenders. For all settings, \ours achieves optimality nearly identical to the optimal DP and ILP (when DP and ILP can complete the computation). ILP starts to fail as the number of defenders reaches $5$ and the number of events exceeds $300$.}
    \label{fig:quality}
\end{figure}

To test and compare the general performance of these methods, we set the defending task on $S^1$ (i.e., a circular boundary with a distance of $2\pi$) with the attack sequence's coming rate $\lambda$ (used in the Poisson process) proportional to $k$, the number of defenders. The results on computational time and interception rate are given in Figs.~\ref{fig:time_cost} and~\ref{fig:quality}.

For 2 to 7 defenders, in Fig.~\ref{fig:time_cost}, we show the computation time of running the algorithms. 
Each data point is an average of 20 runs. 
We remind the readers that both DP and ILP guarantee to produce an optimal solution (if they can finish).
For the two-defender scenario, which is the main focus of \cite{adler2022role}, DP is the most efficient.
However, DP only fully scales for $k=2, 3$ defenders, and start to peter out for $k=4$ defenders due to memory limit (so its running time is not shown for $k>4$).
As can be observed, ILP demonstrates much better scalability in comparison to DP, but fails to find a solution for $k > 5$ defenders. 

As expected, exhaustive defender pairing (\ours) has the least time cost for $k \geq 3$.
Fig.~\ref{fig:quality} shows the corresponding solution qualities of these methods. 
We observe that, while \ours does not guarantee solution optimality, it generated solutions that are virtually the same as the exact methods (DP and ILP). 
Through these empirical evaluations, \ours demonstrates superior scalability with negligible solution optimality loss. 


\subsection{Scaling up the Number of Defenders}
\begin{figure}[h!]
    \centering
    \includegraphics[width=\linewidth]{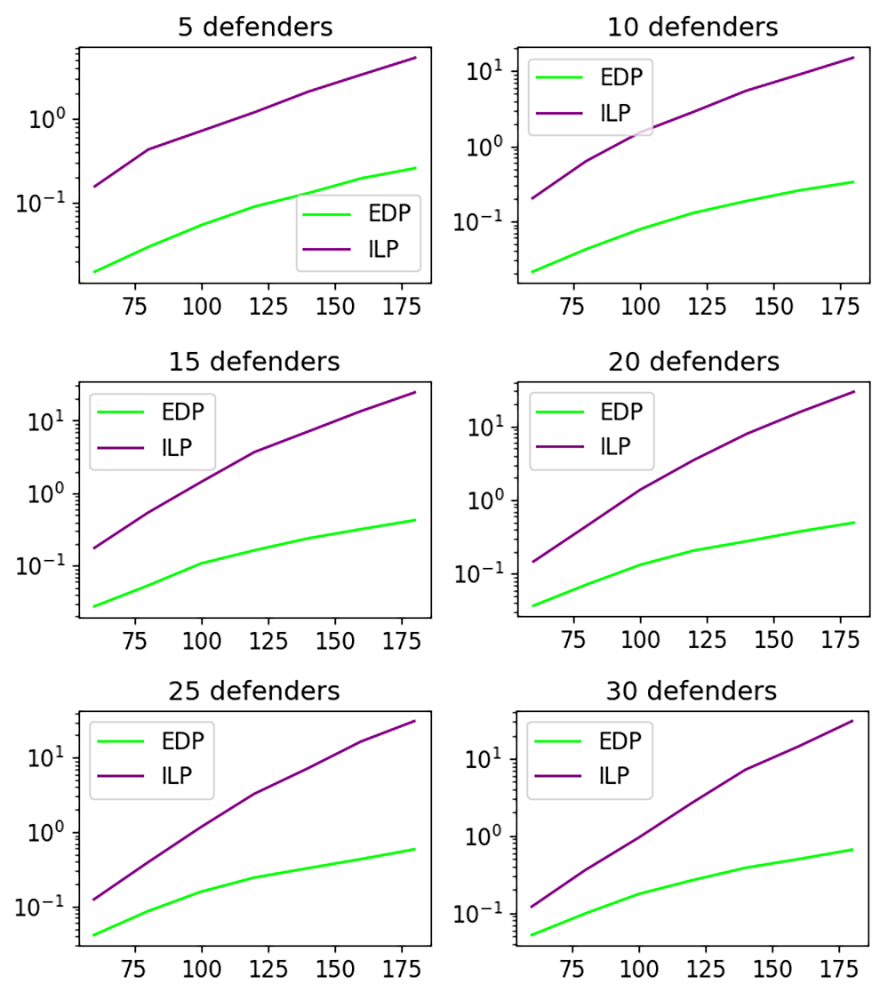}
    \caption{Computation time in seconds for ILP and EDP algorithms for $5$ to $30$ defenders and up to $200$ attack events (Limit the number of events because as shown in Fig.~\ref{fig:time_cost}, more events will effectively cripple ILP). \ours consistently demonstrates 1-2 magnitudes faster computation time compared with ILP.}
    \label{fig:def_time_cost}
\end{figure}

\begin{figure}[h!]
    \centering
    \includegraphics[width=\linewidth]{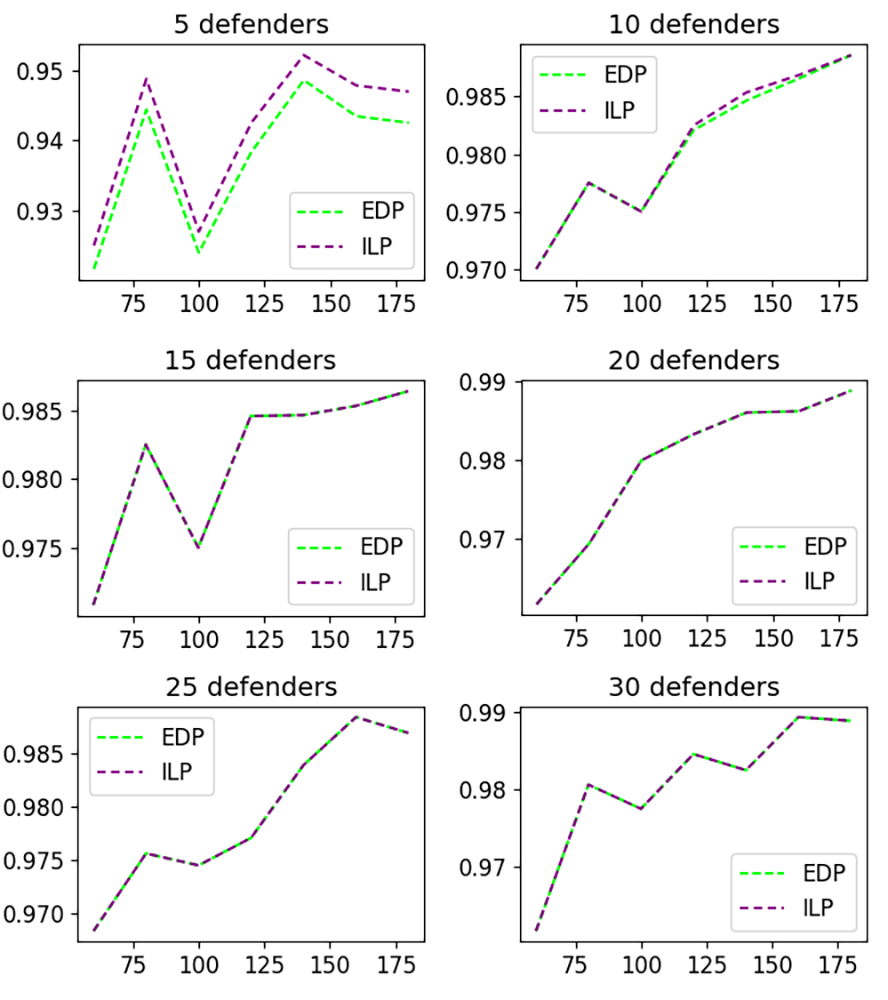}
    \caption{Solution quality (interception rate) for the ILP and EDP for 5 to 30 defenders. It is straightforward to observe that \ours achieves defending quality nearly identical to the optimal ILP solution (while ILP is still scalable).}
    \label{fig:def_quality}
\end{figure}

The number of defenders appears in the index term of the time complexity of the DP algorithm, which limits it from scaling up the number of defenders. Hence, we compare the performance of ILP and the local search algorithm when pushing the number of defenders up to dozens. With the main goal being the evaluation of optimality, we limit the attack events to $200$ so that the ILP method can return the optimal solution in $60$ seconds. At the same time, to make the problem more challenging, $\lambda$ is increased to $2\cdot k$ from $k$ so that not all attacks can be intercepted. Fig.~\ref{fig:def_time_cost} and Fig. ~\ref{fig:def_quality} show the corresponding computation time and interception rate, each data point is the average of 20 runs.

As can be observed, EDP, due to its quadratic dependence on the number of defenders, consistently and significantly outperforms the ILP method in terms of computation time. At the same time, it yields virtually the same level of optimality as ILP, which guarantees that the result is optimal. 

\subsection{Impact of Defender Heterogeneity}
\begin{figure}[h!]
    \centering
    \includegraphics[width=0.9\linewidth]{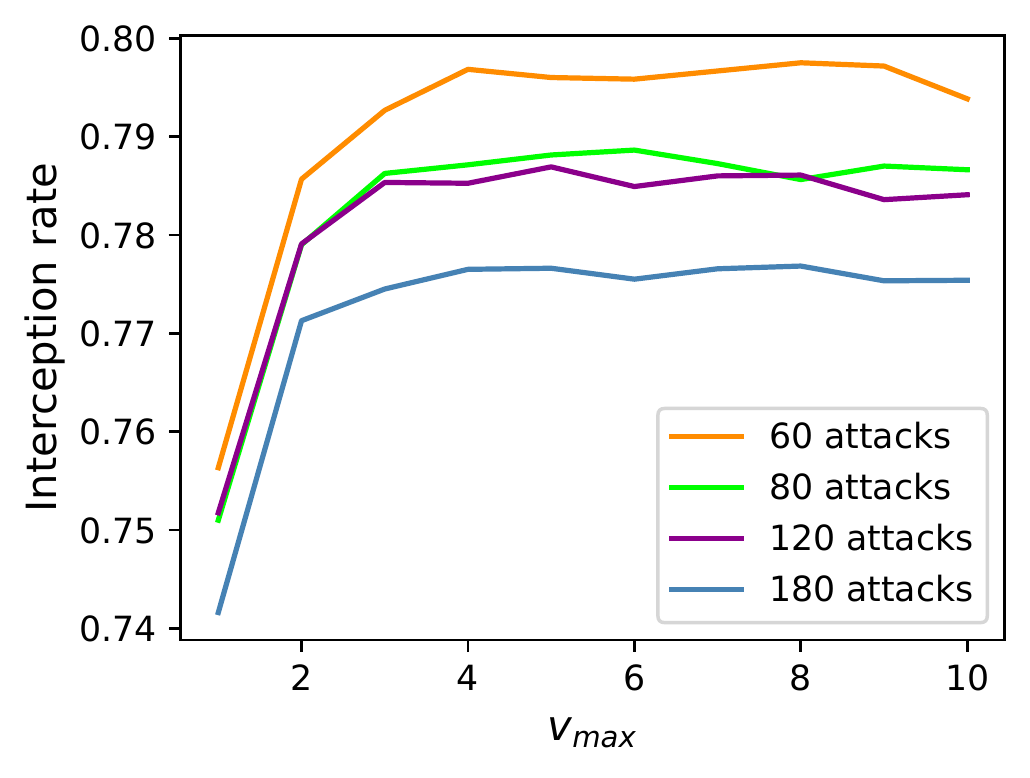}
    \caption{Solution quality (interception rate) for a different level of defender speeds diversity. It can be observed that the major difference happens as $v_{max}:v_{min}$ increases from $1$ to up to $4$. After that, the increase of interception rate no longer shows improvements.}
    \label{fig:heterogeneity}
    \vspace{-2mm}
\end{figure}

To explore the impact of the diversity of defender speeds on interception rate, we focus on a $5$-defender setting and to make the difference more visible, $\lambda$ is set to be $5\cdot k = 25$ so that the interception rate no more than 80\%. 
Here, the defender speed diversity is tested by setting the $v_{min}$ as 1, and $v_{max}$ from 1 (uniform speed) to 10, and then normalizing the defender speed to make them sum up to $15$ (i.e., with an average of $3.0$). 
Each data point is based on 100 runs. The result is summarized in Fig.~\ref{fig:heterogeneity}, from which we can observe that uniform speed $v_{max} = 1$ has the least interception rate, while the interception rate gets into its maximum after $v_{max}:v_{min} = 3$ or $4$. This suggests that heterogeneity can significantly enhance the interception rate, warrant the effort going into the current study (and previous study of heterogeneous defender setups). At the same time, as heterogeneity continues to increase, the effect becomes negligible. This also follows intuition: when the capability of the defenders varies too much, they can no longer effectively collaboratively intercept attacks. 

\subsection{Impact of Attack Rate $\lambda$}
\begin{figure}[h!]
    \vspace{-2mm}
    \centering
    \includegraphics[width=0.9\linewidth]{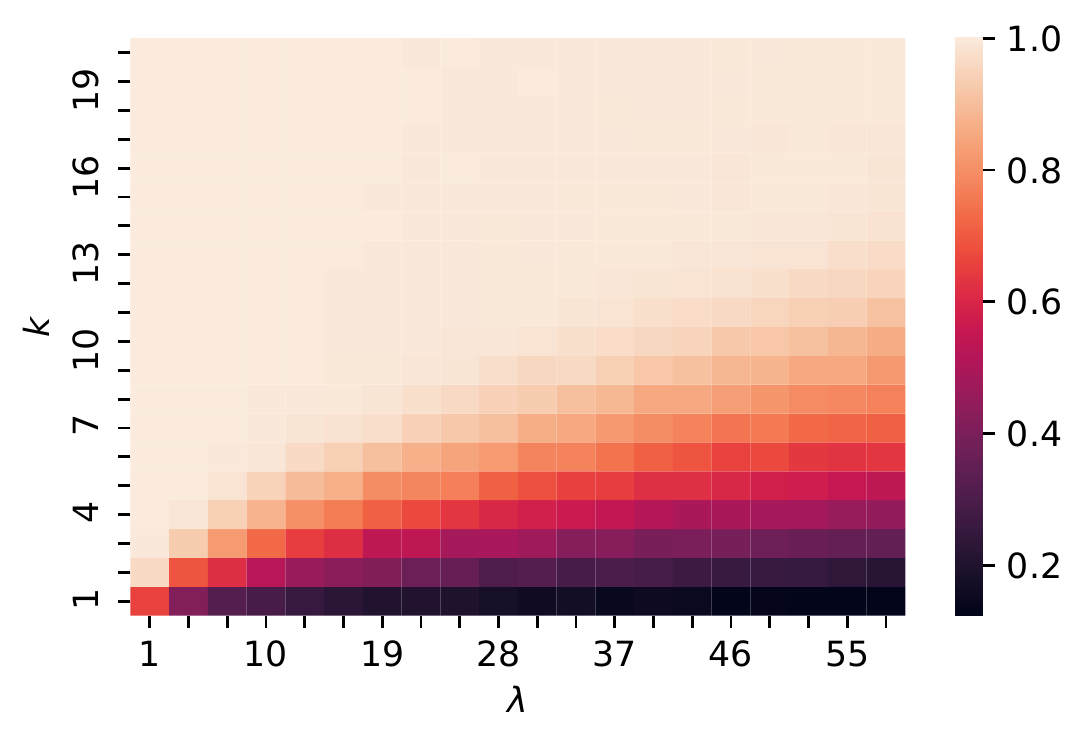}
    \vspace{-2mm}
    \caption{Interception rate with different $\lambda$ (x-axis) and $k$ (y-axis, number of defenders). As one might expect, as attacks happens more frequently, holding other factors unchanged, the interception rate decreases. Similarly, holding other factors unchanged, increasing the number of defenders leads to increased interception rates.}
    \label{fig:lambda}
    \vspace{-2mm}
\end{figure}

With the increase of $\lambda$ in the Poisson process for the attack sequence, 
there will be fewer attacks intercepted, 
and more defenders will be required to reach the same interception rate. 
In this regard, we test the interception rate computed using \ours for $1$ to $20$ defenders and different $\lambda$ from $1.0$ to $60.0$. 
The number of attacks is set to be $400$, and each data point represents an average of 20 runs.
Fig.~\ref{fig:lambda} shows the resulting capture rate with respect to the attack rate $\lambda$ as a heat map for a different number of defenders $k$, which, unsurprisingly, suggests a gradual change as one might expect. Nevertheless, such a graph can serve as a quick reference for selecting an appropriate number defenders given an expected attack rate.

\subsection{Impact of Boundary Topology}
\begin{figure}[h!]
\vspace{2mm}
    \centering
    \includegraphics[width=\linewidth]{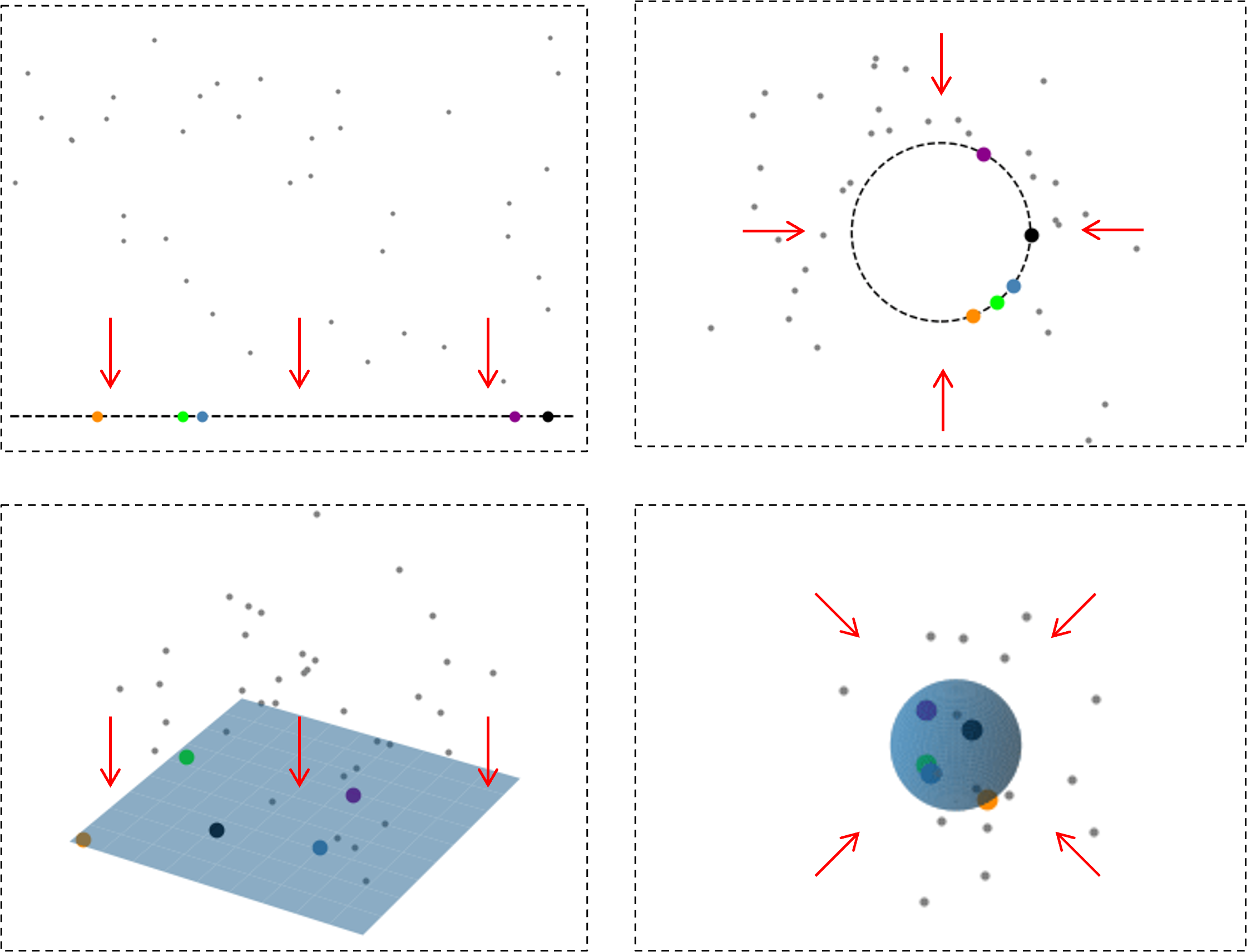}
    \caption{\prob with boundary topologies $I$ (line segment), $S^1$ (circle), $I^2$ (unit square) and $S^2$ (sphere). The red arrows illustrate attack directions. The pictures are captured from animations of actual simulations, available at \url{https://youtu.be/x0mQD\_7RhKI}.}
    \label{fig:topology}
    \vspace{-4mm}
\end{figure}

\begin{figure}[h!]
\vspace{1mm}
    \centering
    \includegraphics[width=\linewidth]{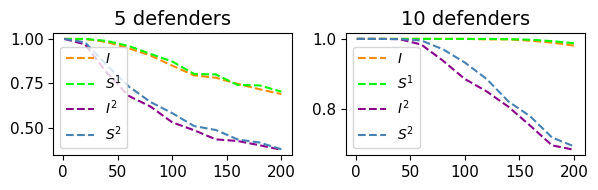}
    \caption{Interception rate for $I$, $S^1 (\text{with length 1})$, $I^2$, $S^2 (\text{with surface area 1})$ with different $\lambda$ from $1$ to $200$. Degradation as the number of attacks increases is as expected.}
    \label{fig:geo}
\end{figure}

\ours applies to \prob with arbitrary boundary topology; we evaluated \ours on $I=[0, 1]$ (unit interval), $S^1$ (circle with a total circumference shrinked to $1$), $I^2$ (unit square), and $S^2$ (sphere with surface area of $1$) (see Fig.~\ref{fig:topology}) with $k = 5$, $v_{max} = 5$ and $\lambda$ from 1 to 200. 
The results are shown in Fig.~\ref{fig:geo}, where each data point is the average of 20 runs. 
We observe that $I$ and $S^1$ have similar difficulty to defend, with $I$ being a bit more challenging due to having boundaries. The same is observed for $I^2$ and $S^2$.

\subsection{Infinite Attack Stream with Finite Look-Ahead Horizon}
\begin{figure}
    \vspace{-1mm}
    \centering
    \includegraphics[width=0.98\linewidth]{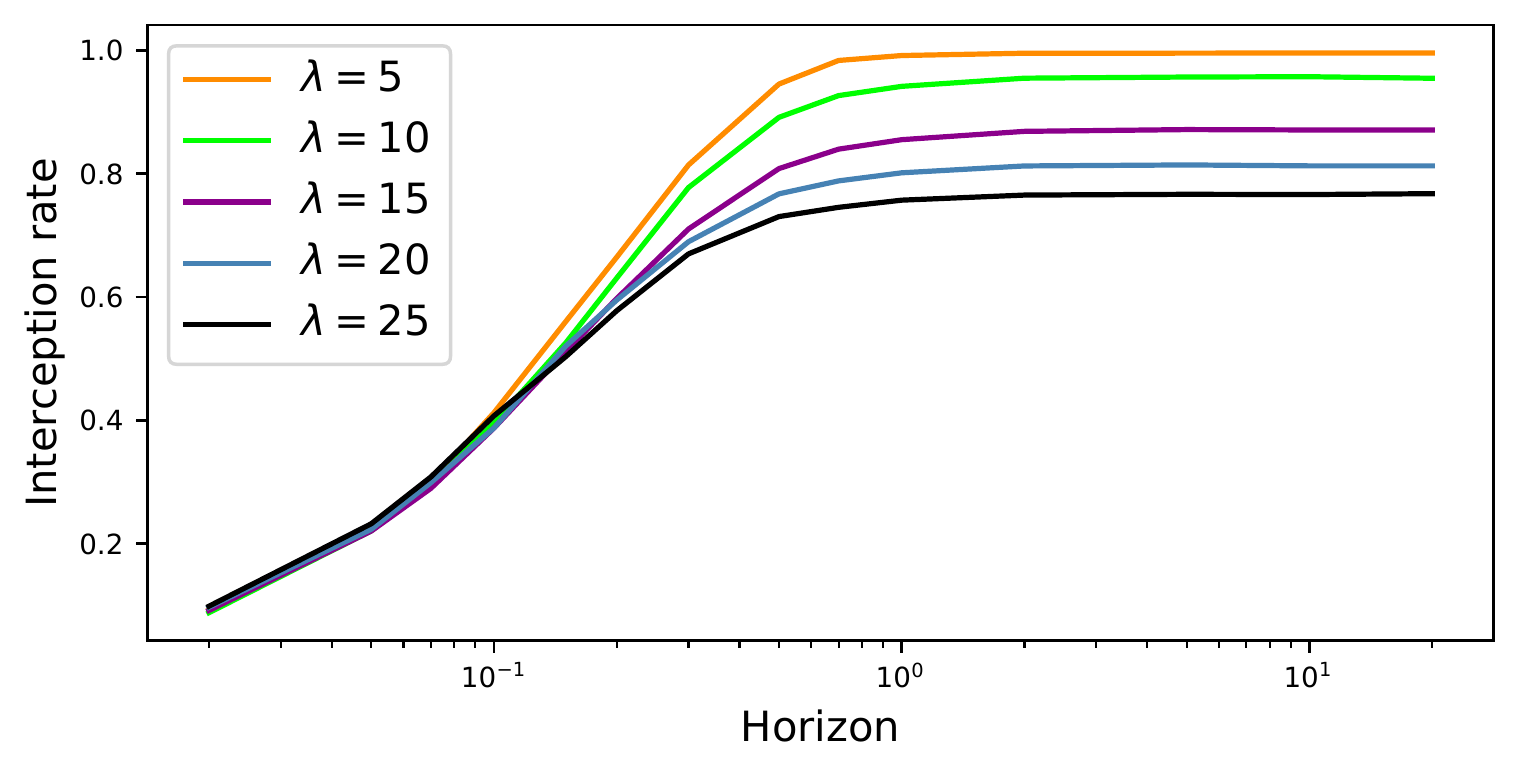}
    \vspace{-2mm}
    \caption{Interception rate with respect to different horizons (logarithmic scale). It can be observed that longer horizon allows better planning.}
    \label{fig:horizon}
    \vspace{-3mm}
\end{figure}

Alg.~\ref{alg:horizon} describes an online \ours algorithm for handling the finite horizon version of the problem where the defenders only have the information about attacks in the next $T$ time frame.
Here, we present some simulation studies using the 5-defender and 400 attacks setup on an $S^1$ boundary with $v_{max} = 5$. Fig.~\ref{fig:horizon} gives the interception rate with respect to varying defender horizons. Each data point is an average of 20 runs. The interception rate converges to around $1$ with a fairly short horizon, indicating that a long look-ahead horizon may not be necessary for effectively intercepting attacks.

\begin{figure}[h!]
    \vspace{1mm}
    \centering
    \includegraphics[width=\linewidth]{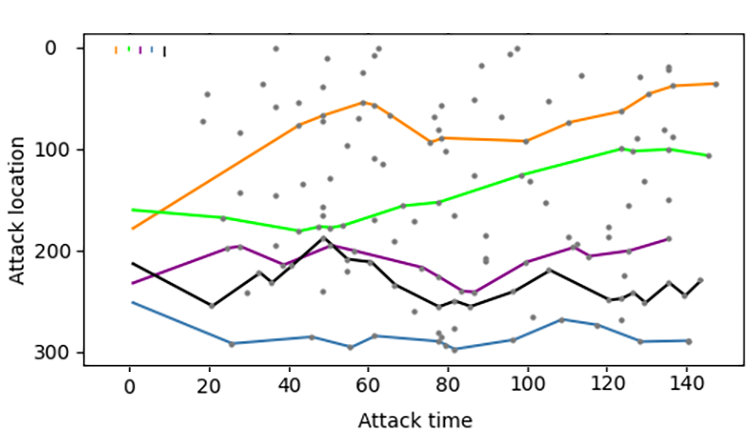}
    \includegraphics[width=\linewidth]{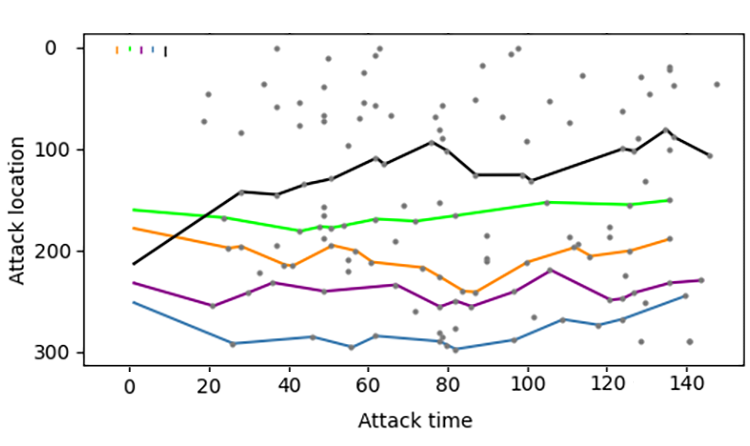}
    \includegraphics[width=\linewidth]{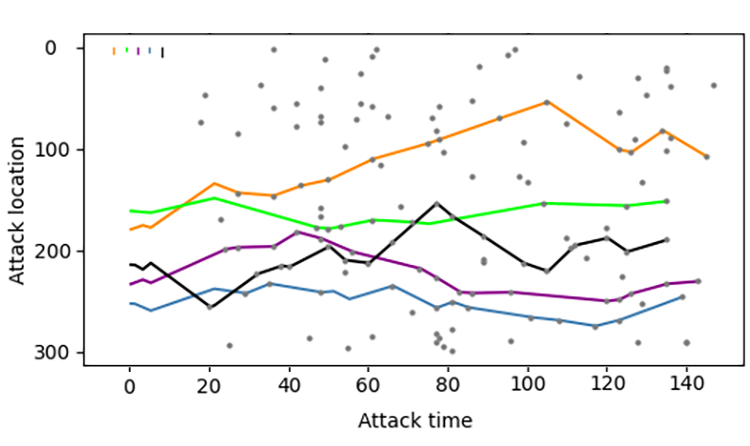}
    \caption{The results computed by ILP, \ours, and online \ours with a horizon of 60, for a sequence of 140 attacks ($x$-axis is the location, $y$-axis is the time of attacks). 
    The number of captured attack events is 71, 70, and 67, respectively.}
    \label{fig:solution_examples}
    \vspace{-4mm}
\end{figure}

Lastly, in Fig.~\ref{fig:solution_examples}, we provide the interception trajectories of a typical run for ILP,  \ours, and online \ours. 

%% file: tex/conclusion.tex
This paper studied the heterogeneous perimeter defense problem, as formulated in \cite{adler2022role}, with the aim of
significantly boosting the scalability to work on many defenders while achieving high levels of solution optimality.
Toward achieving this goal, we developed an exact integer programming-based algorithm with better scalability than the original dynamic programming (DP) based algorithm from \cite{adler2022role}. We then further developed a fast and highly-effective heuristic based on the pairwise application of DP for computing near-optimal solutions, scaling to dozens of defenders.
This pairwise application of DP is of independent algorithmic interest. 
In addition, we extend our solution to a more realistic online version in which only attackers within a finite time window are visible. 
The algorithms are thoroughly evaluated on a diverse set of boundary topologies including circles, intervals, spheres, and squares. 
The evaluation not only confirms that \ours is highly effective but clearly demonstrates the benefit of employing heterogeneous defenders.